%% file: preprint.tex
\documentclass{article} 
\usepackage{lmrl2025_workshop,times}
\lmrlfinalcopy
\input{math_commands.tex}

\usepackage{hyperref}
\usepackage{url}
\usepackage{graphicx}

\title{Implicit Neural Representations of Molecular Vector-Valued Functions}

\author{Jirka Lhotka  \\
LTS2 \\
École Polytechnique Fédérale de Lausanne \\
Lausanne, Switzerland \\
\And
Daniel Probst \\
LTS2 \\
École Polytechnique Fédérale de Lausanne \\
Lausanne, Switzerland \\
\\
Bioinformatics Group \\
Wageningen University \& Research \\
Wageningen, The Netherlands \\
\texttt{daniel.probst@wur.nl} \\
}

\begin{document}

\maketitle

\section{Introduction}
Molecules have various computational representations, including numerical descriptors, strings, graphs, point clouds, and surfaces. Each representation method enables the application of various machine learning methodologies from linear regression to graph neural networks paired with large language models~\citep{serraLinearRegressionComputational2001,heExplainingGraphNeural2024}.

To complement existing representations, we introduce the representation of molecules through vector-valued functions, or $n$-dimensional vector fields, that are parameterized by neural networks, which we denote molecular neural fields. Unlike surface representations, molecular neural fields capture external features and the hydrophobic core of macromolecules such as proteins. Compared to discrete graph or point representations, molecular neural fields are compact, resolution independent and inherently suited for interpolation in spatial and temporal dimensions~\citep{sitzmannDeepVoxelsLearningPersistent2019,lombardiNeuralVolumesLearning2019,jiangLocalImplicitGrid2020}. These properties inherited by molecular neural fields lend themselves to tasks including the generation of molecules based on their desired shape, structure, and composition, and the resolution-independent interpolation between molecular conformations in space and time. Here, we provide a framework and proofs-of-concept for molecular neural fields, namely, the parametrization and superresolution reconstruction of a protein-ligand complex using an auto-decoder architecture and the embedding of molecular volumes in latent space using an auto-encoder architecture.

\begin{figure}[h]
    \centering
    \includegraphics[width=01.0\textwidth]{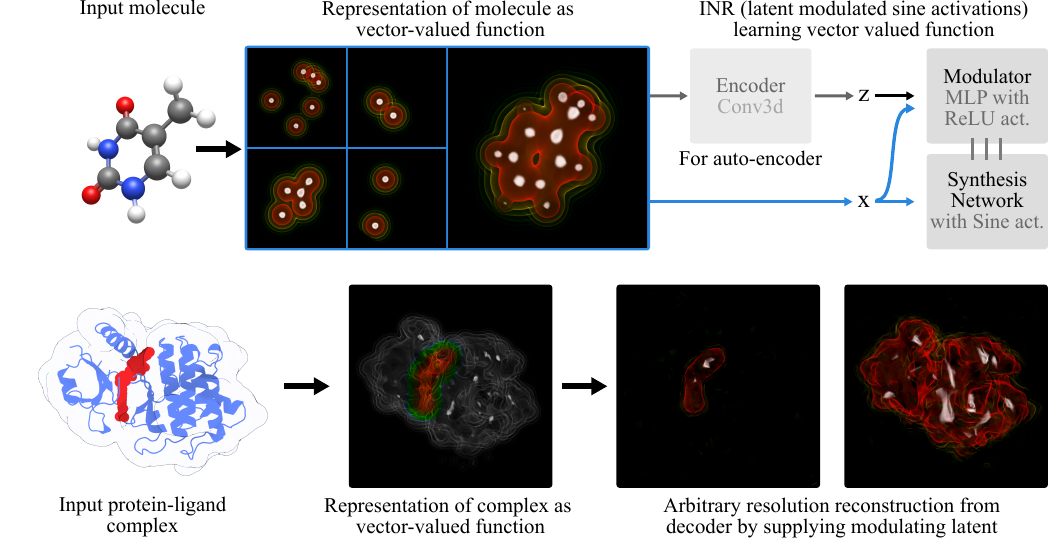}
    \caption{\textbf{Upper row}: Workflow and architecture overview of the presented methodology. For a given input molecule, a vector-valued function is computed and sampled on a 3D grid. Based on these samples, a neural network is trained that parametrizes the molecule using a modulated synthesis network with sine activations introduced by~\citet{mehtaModulatedPeriodicActivations2021}. \textbf{Lower row}: Based on the above workflow and architecture, a protein-ligand complex is encoded and parameterized by a modulated neural network in an auto-decoder setting. The two rightmost images show the reconstruction of the ligand and protein using the learned latent.}
    \label{fig:1}
\end{figure}

\section{Related Work}
Since their introduction by~\citet{sitzmannImplicitNeuralRepresentations2020}, implicit neural networks have been widely used in computer vision for processing images, videos and 3D scenes~\citep{essakineWhereWeStand2024}. More recently,~\citet{mehtaModulatedPeriodicActivations2021} have introduced modulated periodic activations, enabling generalizable high-fidelity implicit neural representations. 

Recent approaches in representation learning of 3-dimensional molecular structures have mainly focused on graph-related methods~\citep{zhouUniMolUniversal3D2022,moon3DGraphContrastive2023}. For the representation of proteins, learning on the molecule's surfaces has proven successful~\citep{gainzaDecipheringInteractionFingerprints2020}. \citep{gainzaDecipheringInteractionFingerprints2020} have used implicit neural networks to generate the surfaces of metal-organic frameworks through signed-distance fields. \citep{kilgourMultiTypePointCloud2024} have introduced multi-type point clouds-representation learning based on an equivariant graph neural network.

\section{Methodology}
A Python library was implemented to sample a vector-valued function $f(x,y,z)$, mapping from $\mathbb{R}^3$ (molecular structure) to $\mathbb{R}^d$ (molecular properties). Each dimension in $\mathbb{R}^d$ represents one or more atom types (e.g. hydrogens, carbons, or oxygens) based on an atomic radius (van der Waals radius as default) or an atomistic property (e.g. electronegativity or Awigner-Seitz electron density). Values in $\mathbb{R}^d$ are computed by the sum of radial basis functions scaled by a radius or property centered at the origin of each atom with given type(s) in the molecule. Section \ref{sec:eq} describes the radial basis function used. The result of this approach is shown in the second step of the upper row of Figure \ref{fig:1}. The neural network architecture is based on an implementation of the modulated periodic activation-based architecture by~\citet{mehtaModulatedPeriodicActivations2021}, extended with the ability to process $d$-dimensional data in order to parameterize vector-valued functions and a 3D convolutional encoder. In this study, the input vector-valued functions of the molecules were uniformly sampled using a $32\times32\times32$ grid. In cases where only SMILES were available, conformers were generated using RDKit.

\section{Results and Conclusion}
For our initial experiment, we trained the proposed architecture as an auto-decoder on a type 2 inhibitor binding to unphosphorylated interleukin-1 receptor-associated kinase 4 (PDB~ID~\texttt{6EGA}). The complex was learned by sampling a $32\times32\times32$ grid of its vector-valued function mapping $\mathbb{R}^3\rightarrow\mathbb{R}^6$ (hydrogens, carbons, and Awigner-Seitz electron density of the ligand and protein) and successfully reconstructed using the conditioning latent with a $\text{PSNR}$ (peak signal-to-noise ratio) of 38.5. Following the super-resolution approach from computer vision, we generated an up-scaled version with a resolution of $128\times128\times128$ ($\text{PSNR}=27.2$) compared to a ground truth sampled at the same higher resolution~\citep{wangDeepLearningImage2021}. A visual comparison of the results is shown in Appendix Figure \ref{fig:2}. Next, we parameterized multiple conformers of a molecular complex, the solution structure of a dimeric lactose DNA-binding domain complexed with a DNA sequence (PDB~ID~\texttt{1OSL}), using a single modulated network trained as an auto-encoder (see architecture in Figure~\ref{fig:1}). An animation of the interpolation through latent space by linearly interpolating between latents can be found in the GitHub repository linked below. Finally, we used the proposed architecture as an auto-encoder trained on the FreeSolv ($n=642$) data set. We sampled a vector-valued function mapping $\mathbb{R}^3\rightarrow\mathbb{R}^{10}$ (all occurring atom types and Awigner-Seitz electron density). The reconstruction quality is similar to the up-scaling experiment with an average $\text{PSNR}$ of $28.7$. Upon visual inspection of the latent space, we note that the latents represent the shape, size, and basic physicochemical properties of encoded molecules (Appendix Figure \ref{fig:3}). In Appendix Figure \ref{fig:4}, we show that the model can also smoothly interpolate between randomly chosen molecules when trained on a larger data set.

Molecular neural fields have a vast potential for applications in machine learning for life sciences. Learning directly on electron density maps from X-ray crystallographic experiments or Cryo-EM density maps are near-future applications. Beyond that, the representation allows for the simple integration of organic and non-organic molecular information and the parametrization of physicochemical valid navigation between conformers. While challenges remain, we believe this work to be valuable to the broader machine learning community in life sciences. All code, models, and data are available at \url{https://github.com/daenuprobst/minf}.

\bibliography{preprint}
\bibliographystyle{iclr2025_conference}
\clearpage
\appendix
\section{Appendix}

\begin{figure}[h]
    \centering
    \includegraphics[width=0.9\textwidth]{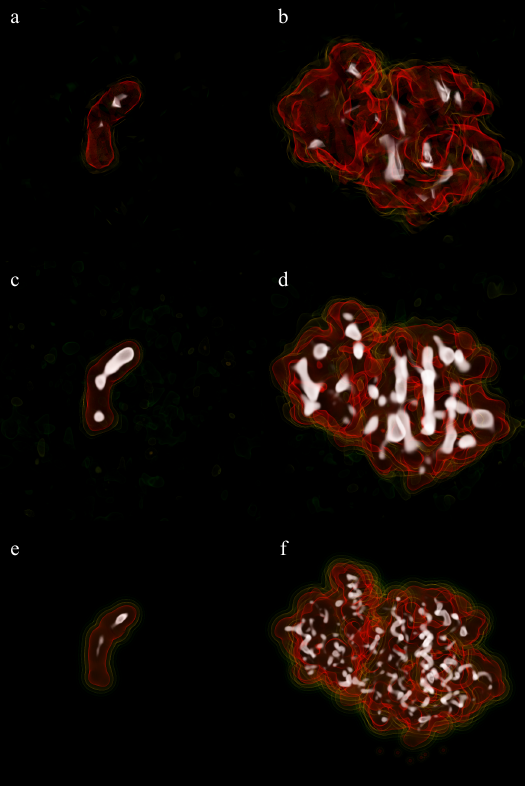}
    \caption{Super-resolution decoding of protein-ligand complex (separated channels). The top row (a, b) shows a density plot with a resolution $32\times32\times32$. The middle row (c, d) shows the up-scaled version with a resolution of $128\times128\times128$. The bottom row (e, f) shows the ground truth at a resolution of $128\times128\times128$. The peak signal-to-noise ratio of the up-scaled version and the ground truth is 75.7.}
    \label{fig:2}
\end{figure}

\begin{figure}[h]
    \centering
    \includegraphics[width=1.0\textwidth]{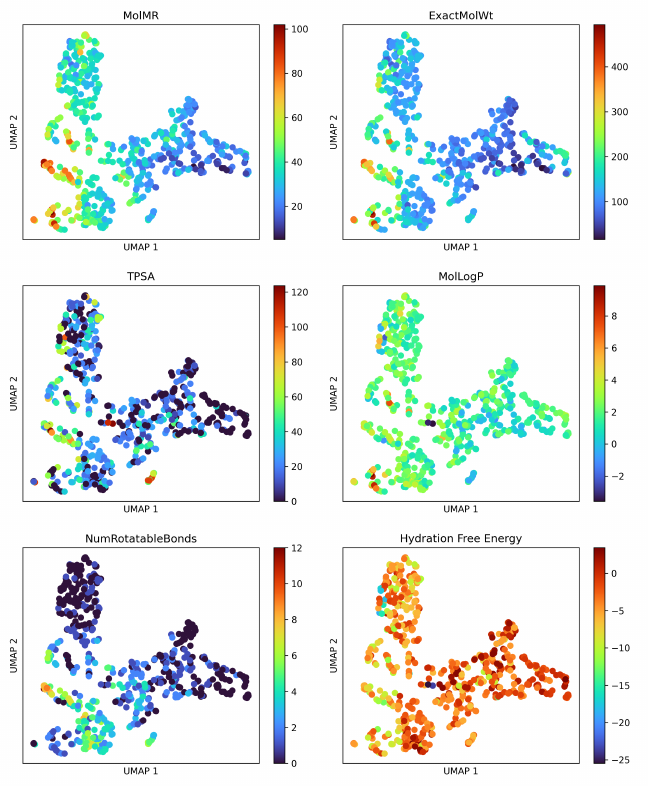}
    \caption{Embedding of molecular neural field latents of the FreeSolv data set. The scatter plots are colored by the Wildman--Crippen molar revlectivity value (MolMR), the exact molecular weight (ExactMolWt), the topological polar surface area (TPSA), the Wildman--Crippen LogP value (MolLogP), the number of rotatable bonds (NumRotatableBonds), and the hydration free energy. The hydration free energy was provided with the FreeSolv data set, all other values were calculated using RDKit. On visual inspection, the plots show that the latent space is meaningful in terms of molecular structure and shape.}
    \label{fig:3}
\end{figure}

\begin{figure}[h]
    \centering
    \includegraphics[width=1.0\textwidth]{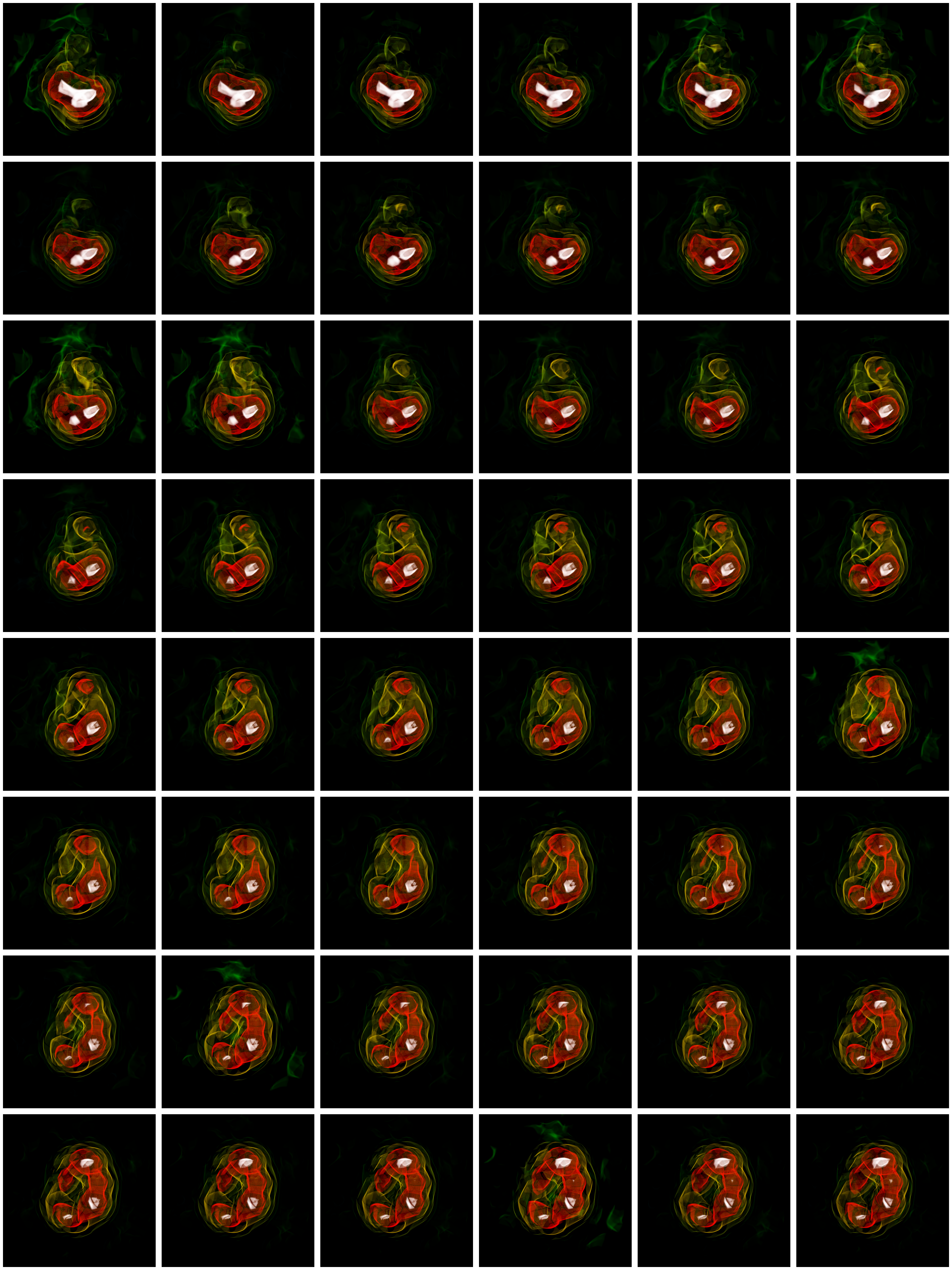}
    \caption{Smoothly interpolating between two molecules parametrized by the auto-encoder trained on the FreeSolv data set.}
    \label{fig:4}
\end{figure}

\clearpage
\subsection{Equations}
\label{sec:eq}

The vector-valued function $f$ mapping from $\mathbb{R}^3$ to $\mathbb{R}^1$ for a given atom type $t$ is sampled at point $(x,y,z)$ using the following function:

\begin{equation} 
f(x,y,z)= \sum_{a_i\in M_t}{\exp\left(-\beta\left(\frac{d(a_i, (x,y,z))}{r_a^2}-1\right)\right)}
\end{equation}

Where $a_i$ are the coordinates for atom $a_i\in M_t$, $\beta$ a user-chosen scaling factor, $d(a_i, (x,y,z))$ the squared Euclidean distance between $a_i$ and $(x,y,z)$, and $r_a$ the chosen radius property for atom $a_i$. As default, the radius property for atom field calculations is the Van der Waals radius.

\end{document}

%% file: math_commands.tex
\usepackage{amsmath,amsfonts,bm}

\def\eqref#1{equation~\ref{#1}}

\def\1{\bm{1}}

\DeclareMathAlphabet{\mathsfit}{\encodingdefault}{\sfdefault}{m}{sl}
\SetMathAlphabet{\mathsfit}{bold}{\encodingdefault}{\sfdefault}{bx}{n}

